\pdfoutput=1
\documentclass[11pt]{article}

\usepackage{acl}

\usepackage{times}
\usepackage{latexsym}

\usepackage[T1]{fontenc}
\usepackage[utf8]{inputenc}

\usepackage{microtype}

\usepackage{makecell}
\usepackage{multirow}
\usepackage{amsmath}
\usepackage{amssymb}
\usepackage{enumitem}
\usepackage{color}
\usepackage{graphicx}
\usepackage{comment}

\usepackage{geometry}
\usepackage{algorithm2e}

\title{uChecker: Masked Pretrained Language Models as Unsupervised \\Chinese Spelling Checkers}

\author{Piji Li\\
College of Computer Science and Technology,\\
Nanjing University of Aeronautics and Astronautics\\
MIIT Key Laboratory of Pattern Analysis and Machine Intelligence\\
Nanjing, Jiangsu, China\\
\texttt{pjli@nuaa.edu.cn}
}

\begin{document}
	\maketitle
	\begin{abstract}
		The task of Chinese Spelling Check (CSC) is aiming to detect and correct spelling errors that can be found in the text. While manually annotating a high-quality dataset is expensive and time-consuming, thus the scale of the training dataset is usually very small (e.g., SIGHAN15\footnote{\url{http://ir.itc.ntnu.edu.tw/lre/sighan8csc.html}} only contains 2339 samples for training), therefore supervised-learning based models usually suffer the data sparsity limitation and over-fitting issue, especially in the era of big language models. In this paper, we are dedicated to investigating the \textbf{unsupervised} paradigm to address the CSC problem and we propose a framework named \textbf{uChecker} to conduct unsupervised spelling error detection and correction. Masked pretrained language models such as BERT are introduced as the backbone model considering their powerful language diagnosis capability. Benefiting from the various and flexible MASKing operations, we propose a Confusionset-guided masking strategy to fine-train the masked language model to further improve the performance of unsupervised detection and correction. Experimental results on standard datasets demonstrate the effectiveness of our proposed model uChecker in terms of character-level and sentence-level Accuracy, Precision, Recall, and F1-Measure on tasks of spelling error detection and correction respectively.

	\end{abstract}
	
	\section{Introduction}
	\begin{figure}[t!]
		\centering
		\includegraphics[width=0.95\columnwidth,height=0.33\columnwidth]{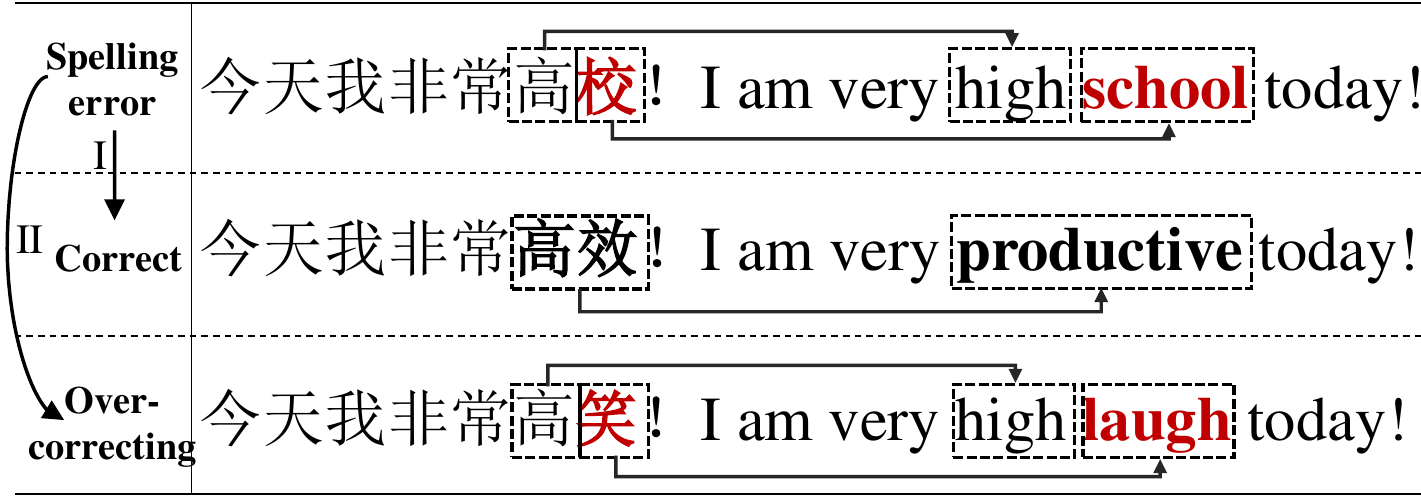}
		\caption{Illustration for the task of Chinese spelling check (operation path I) as well as the over-fitting phenomenon existing in the current supervised learning based models (operation path II).}
		\label{fig:front}
	\end{figure}
	
	Chinese Spelling Check (CSC) is a crucial and essential task in the area of natural language processing. It aims to detect and correct spelling errors in the Chinese text~\cite{chang1995new,DBLP:journals/corr/abs-2005-06600}. 
	Generally, sequence translation~\cite{DBLP:conf/emnlp/WangSLHZ18,DBLP:journals/corr/abs-1807-01270,DBLP:conf/acl/WangTZ19,DBLP:conf/ijcnlp/WangKKK20,DBLP:conf/acl/KanekoMKSI20} and sequence tagging~\cite{DBLP:conf/bea/OmelianchukACS20,liang-etal-2020-bert,DBLP:conf/emnlp/MallinsonSMG20,DBLP:conf/acl/ParnowLZ21} are the two most typical technical paradigms to tackle the problem. Benefiting from the development of pretraining techniques, many researchers fine-tune the pretrained language models such as BERT~\cite{DBLP:conf/naacl/DevlinCLT19} on the task of CSC and obtain encouraging performance~\cite{DBLP:conf/naacl/ZhaoWSJL19,DBLP:conf/aclnut/HongYHLL19,DBLP:conf/acl/ZhangHLL20,DBLP:conf/acl/LiuYYZW20,DBLP:conf/acl/LiZZH20,DBLP:conf/acl/HuangLJZCWX20,DBLP:conf/acl/GuoNWZX21,DBLP:conf/acl/ZhangPZWHSWW21,DBLP:conf/acl/Li020,dai-etal-2022-whole}. Meanwhile, it should be emphasized that almost all of the above mentioned models are trained via the \textbf{supervised learning} paradigm. 
	
	However, during the investigating stage about those newly typical state-of-the-art models, we observe some spiny and serious phenomenons: (1) Occasionally those models may generate some special \textbf{over-correcting} results. As shown in Figure~\ref{fig:front}, operation path \textbf{I} is the regular spelling error detection and correction path, while operation path \textbf{II} is also observable in the inference stage where the models can detect the errors correctly but rectify them using some other error tokens in the correction stage. (2) The spelling error detection and correction performance will drop dramatically when those models did not see the spelling error cases in the training dataset or the text are from different genres and domains. This issue tells us that the \textbf{generalization capability} of those models are limited and need to be enhanced.
	
	\begin{table*}[!t]
		\centering
		\small
		\begin{tabular}{l||c|c|c||c|c|c}
			\Xhline{2\arrayrulewidth}
			Corpus & \#Train & \#ErrTrain & AvgLen & \#Test & \#ErrTest & AvgLen \\
			\hline
			SIGHAN13 & 350 &  350  & 49.2 & 1,000 & 996 & 74.1  \\
			\hline
			SIGHAN14 & 6,526 & 3,432 & 49.7 & 1,062 & 529 & 50.1  \\
			\hline
			SIGHAN15 & 3,174 & 2,339 & 30.0 & 1,100 & 550 & 30.5 \\
			\hline
			\Xhline{2\arrayrulewidth}
		\end{tabular}
		\caption{Statistics of the SIGHAN series datasets.}
		\label{tab:datasets}
	\end{table*}
	
	Then what are the causes of these phenomenons? Since some of the models are already strong enough (which are constructed based on big pretrained models), then we shift our eyes to the data perspective. In real practical scenarios, natural human-labeled spelling error corpus are difficult and expensive to obtain. Although some works such as \citet{DBLP:conf/emnlp/WangSLHZ18} employ OCR and ASR based techniques to automatically synthetic the paired samples by replacing the correct tokens using visually 
	or phonologically similar characters, obviously, the constructed data is unrealistic and far from the real and objective scenarios. Therefore, actually, the scale of the typical corpus for the task of Chinese spelling check is very small. Considering that almost all the research works have used SIGHAN 
	series datasets \cite{DBLP:conf/acl-sighan/TsengLCC15} to train and evaluate their algorithms, we conduct counting on those three corpora, and the statistics results are shown in Table~\ref{tab:datasets}. From the results we can observe that there are only 2k$\sim$3k sentences with spelling errors in the training dataset and really far from the practical requirements. 
	
	Thus, sticking to train the supervised learning models based on those scale-limited resources might not be a wise direction. Therefore, in this paper, we are dedicated to exploring \textbf{unsupervised} frameworks to conduct  Chinese spelling error detection and correction. Fortunately, masked pretrained language models such as BERT \cite{DBLP:conf/naacl/DevlinCLT19}, RoBERTa~\cite{DBLP:journals/corr/abs-1907-11692}, ELECTRA~\cite{DBLP:conf/iclr/ClarkLLM20}, etc. can satisfy the needs of detecting and correcting spelling errors in an unsupervised manner. First, the masked training strategy is naturally a convenient and perfect shortcut for us to conduct token-grained detection and correction. For example, we can mask any token and predict it based on the bi-directional context to see if the current token is appropriate or not. Second, the pretrained language models are usually trained using large-scale corpora, thus the language diagnosis capability is very strong. Intuitively, these models can also guarantee the generalization capability considering the corpora may contain text from a wide range of domains and genres.
	
	Therefore, based on the masked pretrained language models, we propose a framework named \textbf{uChecker} to conduct unsupervised spelling error detection and correction respectively. uChecker is a two-stage framework and it will detect the text token-by-token first and then correct the abnormal tokens. Models such as BERT are introduced as the backbone model. Inspired by the previous works~\cite{DBLP:conf/acl/WangTZ19,DBLP:conf/acl/LiuYYZW20}, benefiting from the various and flexible masking operations, we also introduce a confusionset-guided masking strategy to fine-train the masked language model to further improve the performance of unsupervised detection and correction. Though uChecker is a two-stage framework, we design an elegant method to let the information pass BERT only once to guarantee the time efficiency. Moreover, interestingly, in unsupervised settings, we experimentally find the performance of error detection is crucial to the global and general performance. It means that the correction capability of the pretrained language models are strong enough, then the key-point is how to improve the performance of detection. Therefore, in uChecker, we also design several algorithms to improve the performance of seplling error detection. \citet{DBLP:conf/emnlp/YasunagaLL21} employ GPT2-like models to conduct unsupervised English grammatical error correction which also verifies the feasible of our direction.   
	
	In summary, our contributions are as follows:
	\begin{itemize}[topsep=0pt]
		\setlength\itemsep{-0.5em}
		\item We propose an unsupervised framework named \textbf{uChecker} to conduct Chinese spelling error detection and correction.
		\item Benefiting from flexible masking operations, we introduce s confusionset-guided masking strategy to fine-train BERT to further improve the performance.
		\item We experimentally find that error detection is crucial to the global SCS performance. Therefore, we also design some algorithms to improve the capability of error detection.  
		\item Extensive experiments on several benchmark datasets demonstrate the effectiveness of the proposed approach. And the results also show that uChecker can even outperform some strong supervised models. 
	\end{itemize}
	
	\section{Background: Masked Language Models}
	
	\begin{figure}[t!]
		\centering
		\includegraphics[width=0.95\columnwidth]{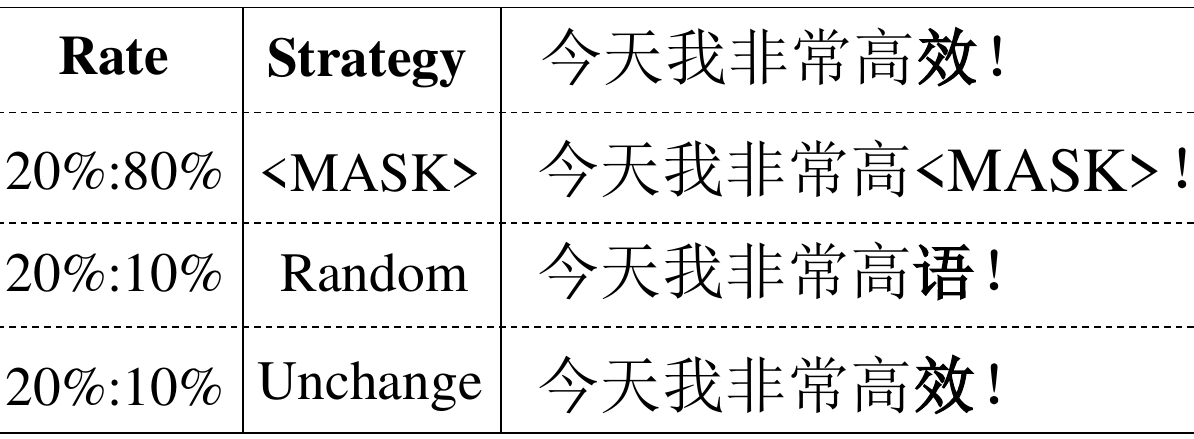}
		\caption{Illustration of the masking strategies of BERT during the pretraining stage.}
		\label{fig:mask}
	\end{figure}
	
	BERT~\cite{DBLP:conf/naacl/DevlinCLT19} is the most typical masked language model, and it is regarded as the backbone model of our proposed framework uChecker, therefore in this section we introduce the technical details of this model, especially the masking strategies.
	
	BERT is constructed based on the model of Transformer \cite{DBLP:conf/nips/VaswaniSPUJGKP17}. After preparing the input samples, an embedding layer and a stack of Transformer layers are followed to conduct the bi-directional semantic modeling. 
	Specifically, for the input, we first obtain the representations by summing the word embeddings with the positional embeddings:
	\begin{equation}
		\mathbf{H}^0_t = \mathbf{E}_{w_t}+\mathbf{E}_{p_t}
	\end{equation}
	where $0$ is the layer index and  $t$ is the state index. $\mathbf{E}_w$ and $\mathbf{E}_p$ are the embedding vectors for tokens and positions, respectively.
	Then the obtained embedding vectors $\mathbf{H}^0$ are fed into several Transformer layers. Multi-head self-attention is used to conduct bidirectional representation learning:
	\begin{equation}
		\begin{split}
			\mathrm{\bf H}^{1}_{t} &= \textsc{Ln}\left(\textsc{Ffn} (\mathrm{\bf H}^{1}_{t}) +\mathrm{\bf H}^1_t \right) \\
			\mathrm{\bf H}^{1}_{t} &= \textsc{Ln}\left(\textsc{Slf-Att} (\mathrm{\bf Q}^{0}_{t}, \mathrm{\bf K}^{0}, \mathrm{\bf V}^{0}) +\mathrm{\bf H}^0_t \right) \\
			\mathrm{\bf Q}^{0} &=  \mathrm{\bf H}^{0} \mathrm{\bf W}^{Q} \\
			\mathrm{\bf K}^{0}, \mathrm{\bf V}^{0} &= \mathrm{\bf H}^{0}\mathrm{\bf W}^{K}, \mathrm{\bf H}^{0} \mathrm{\bf W}^{V}
		\end{split}
		\label{eql:formant_c1}
	\end{equation}
	where \textsc{Slf-Att}($\cdot$), \textsc{Ln}($\cdot$), and \textsc{Ffn}($\cdot$) represent self-attention mechanism, layer normalization, and feed-forward network respectively \cite{DBLP:conf/nips/VaswaniSPUJGKP17}. After $L$ Transformer layers, we obtain the final output representation vectors $\mathrm{\bf H}^{L} \in \mathbb{R}^{T \times d}$, where $T$ is the input sequence length and $d$ is the vector dimension.
	
	The masking strategies used in BERT are shown in Figure~\ref{fig:mask}. There are $20\%$ tokens will be masked, and among them there are $80\%$ tokens are replaced with a special symbol such as $\text{<MASK>}$, and $10\%$ are replaced with a random token, and the left $10\%$ keep unchanged. What should be emphasized here is the \textit{\textbf{random replacing operation}} which plays an crucial role in the following model designs about the unsupervised detection and correction as well as the confusionset-guided fine-training.
	
	Finally, a linear function $g$ with $\operatorname{softmax}$ activation is used to predict the masked token $x_t$ via:
	\begin{equation}
		p \left( x_t | x_{\leq{t-1}},x_{\geq{t+1}}  \right) = \operatorname { softmax } \left( g \left( \mathbf{h}_ { t } \right) \right)
		\label{eq:outprob_bert}
	\end{equation}

	\section{The Proposed uChecker Framework}
	\label{sec:uchecker}
	
	\begin{figure*}[t!]
		\centering
		\includegraphics[width=1.7\columnwidth]{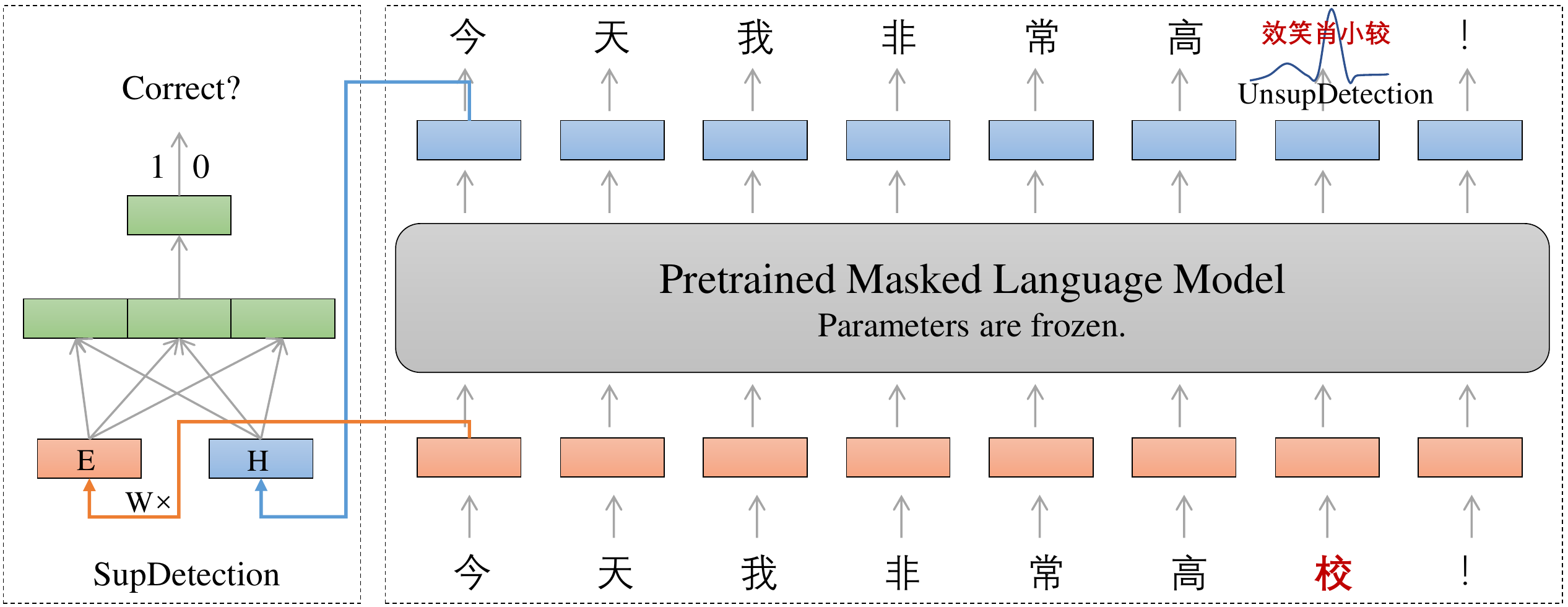}
		\caption{The proposed uChecker framework for unsupervised Chinese spelling error detection and correction.}
		\label{fig:framework}
	\end{figure*}
	
	\subsection{Overview}
	Figure~\ref{fig:framework} depicts the basic components of our proposed framework uChecker. The backbone model is masked language model, say BERT. Note that all the parameters of BERT are frozen. Input is an incorrect sentence $X = (x_1, x_2, \dots, x_T)$ which contains spelling errors, where $x_i$ denotes each token (Chinese character) in the sentence, and $T$ is the length of $X$. The objective of the task Chinese spelling check is to detect and correct all errors in $X$ and obtain a new sentence $Y = (y_1, y_2, \dots, y_{T'})$. Benefiting from the various and flexible masking operations, we introduce the confusionset-guided masking strategy to fine-train BERT to further improve the performance of CSC. We also design several algorithms such as unsupervised detection (UnsupDetection), supervised detection (SupDetection), and Ensemble of UnsupDetection and SupDetection to improve the performance of error detection to further improve the overall performance. 
	
	\subsection{Unsupervised Spelling Error Detection}
	\label{sec:udec}
	
	\begin{figure}[t!]
		\centering
		\includegraphics[width=0.8\columnwidth]{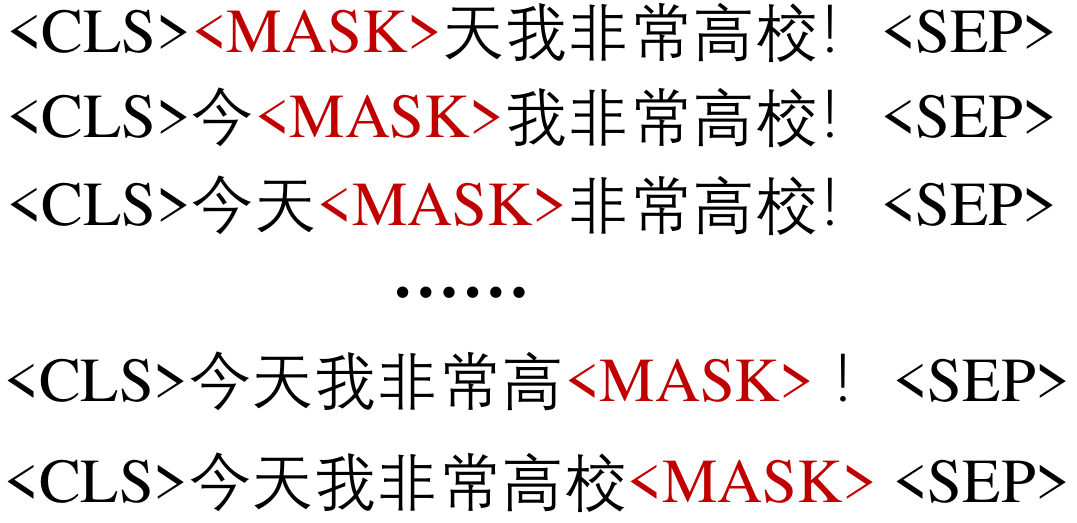}
		\caption{Illustration of the first guess \textbf{diagonal masking strategy} for unsupervised error detection and correction, which is actually \textbf{time-consuming and not necessary}.}
		\label{fig:diagonal}
	\end{figure}
	Given a pretrained BERT model, for a sentence $X = (x_1, x_2, \dots, x_T)$ which need to be checked, the preconceived first guess is to mask the tokens one each time from left to right diagonally, as shown in Figure~\ref{fig:diagonal}, and then input the masked sequence $X'$ into BERT to conduct prediction. Assume token $x_t$ is masked, the predicted distribution at position $t$ is $p(x_t')$, then we can obtain the probability of the corresponding original input token $x_t$ by:   
	\begin{equation}
		p^u_{x_t}=p_{x_t'=x_t}(x_t')
	\end{equation}
	Intuitively, if $x_t$ is just the correct token, then $p^u_{x_t}$ will be very large (say 0.99). Otherwise, error may hide in this position. Therefore, the simple approach to conduct detection is to find a threshold $\theta^u$ and diagnose the results by:
	\begin{eqnarray}
		error_t=\begin{cases}
			1, \ \ \ \text{if} \ \ p^u_{x_t}<\theta^u
			\\
			0, \ \ \ otherwise
		\end{cases}
	\end{eqnarray}
	where $error_t=1$ means that token $x_t$ is not correct in sentence $X$.
	
	However, although diagonal masking strategy is a natural way to conduct token diagnose, considering that for each sentence with length $T$, we need prepare $T$ sequences to feed BERT to conduct prediction, which is a time-consuming procedure with low efficiency, and it is also difficult to be deployed and executed concurrently. Recall the masking strategies shown in Figure~\ref{fig:mask}, besides replacing the tokens with $\text{<MASK>}$ symbol, \textit{BERT also uses \textbf{random tokens} to conduct masking}. Therefore, we do not need to rigidly obey the $\text{<MASK>}$ based masking approach. On the contrary, we can briefly regard the potential error tokens as the random masking strategy and just feed the original input sentence $X$ into the BERT model to conduct probability estimation. And this approach can execute with high concurrency because we can process hundreds or even thousands of sentences in a batch based on the parallel computing capability of GPUs. Moreover, since random masking is feasible, then \textit{how about conduct random masking using the corresponding tokens from the \textbf{confusionset}}? This is the \textbf{inspiration} of confusionset-guided fine-training strategy which will be introduced in the following sections. 
	
	For each sentence $X$, the spelling error detection stage will return a list containing the indices of the wrong tokens $\mathcal{I}^e = [2, 5, i, \dots]$, ranked by the predicted corresponding probability in an ascending order. The order indicates that the most worse token will be corrected firstly.  
	
	\subsection{Unsupervised Spelling Error Correction}
	Given the detected wrong token indices $\mathcal{I}^e$, the unsupervised error correction component then scans the list and chooses the most appropriate tokens from the probability distribution to conduct correction. Specifically, for any index $i \in \mathcal{I}^e$, the predicted distribution is $p(x_i')$, then we can straightforwardly select the token with the largest score as the correct result:
	\begin{equation}
		x_j = \text{argmax}_j \ \ p_{x_i'=x_j}(x_i')
	\end{equation}
	The operation of unsupervised spelling error correction is simple, therefore the correction performance will completely depend on the capability of the pretrained backbone language models.
	
	\paragraph{Confusionset-guided Token Selection}
	Due to the special input methods such as Pinyin and Wubi, many Chinese characters are similar either in phonology or morphology. There are about 76\% of Chinese spelling errors belong to phonological similarity error and 46\% belong to visual similarity error~\cite{DBLP:journals/talip/LiuLTCWL11}. Intuitively, incorporating the Confusionset with the token selection procedure may improve the performance. Therefore, we further build a simple confusionset-guided token selection approach as shown in Algorithm~\ref{alg:token_selection}.
	\RestyleAlgo{ruled}
	\begin{algorithm}[!t]
		\caption{Confusionset-guided Token Correction}\label{alg:token_selection}
		\KwData{Wrong token indices $\mathcal{I}^e$; The predicted distributions for all the positions $\mathcal{P}$; The predefined confusionset $\mathcal{C}$ (hashmap<string, list>).}
		\KwResult{The correct tokens $Y$ for $\mathcal{I}^e$.}
		Y = [] \;
		\For{$i \in \mathcal{I}^e$}{
			$W_i = \text{top\_k}(\mathcal{P}_i)$\;
			\For{$w_i \in W_i$}{
				\If{$w_i \in \mathcal{C}(x_i)$}{
					Y.insert($w_i$)\;
					break\;
				}
				Y.insert($W_i[0]$)\;
			}
		}
	\end{algorithm}
	
	Specifically, for each detected index $i$, we fist fetch the top\_k tokens according to the distribution $\mathcal{P}_i)$. Then if the top\_k tokens are also from the corresponding confusionset, then we get the result. Otherwise, we still select the best predicted result.
	
	\subsection{Self-Supervised Spelling Error Detection}
	Surprisingly and interestingly, during the experiments stage, we find that \textbf{the performance of error detection plays an crucial role} in affecting the global checking performance. Therefore, improving the capability of error detection can benefit the whole system. But, there is a precondition that we cannot adjust the original backbone BERT parameters because unsupervised error correction is the essential component of our uChecker framework, and we do not want to let the BERT parameters collapse to some special areas or domains. So the BERT parameters need to be frozen when designing the error detection strategies.
	
	As shown in Figure~\ref{fig:framework}, after examining the model carefully, we create a \textbf{smart but straightforward self-supervised detection method} to tackle the problem. The basic observation is that, based on the masked language models, the information in the output hidden states $\mathbf{H}$ will be more closer to the true tokens because the model will use them ($\mathbf{H}$) to conduct masked prediction (Eq.~\ref{eq:outprob_bert}). Moreover, the information in the embedding layer $\mathbf{E}$ also contains the token information. Then we assume that for any correct token $x_i$ and error token $x_j$, pair $(\mathbf{e}_i, \mathbf{h}_i)$ for normal token holds a more tight relationship than the pair $(\mathbf{e}_j, \mathbf{h}_j)$ for wrong token, where $\mathbf{e}$ is the learnt token embedding and $\mathbf{h}$ is the output layer of BERT:
	\begin{equation}
		\mathcal{M}(\mathbf{e}_i, \mathbf{h}_i) > \mathcal{M}(\mathbf{e}_j, \mathbf{h}_j)
	\end{equation}
	where $\mathcal{M}$ is metric to represent the interaction relationship, and here we use the following calculations to conduct the  interaction modeling:
	\begin{equation}
		\mathbf{h}^s_i=\mathbf{W}_{s}(\mathbf{e}_i';\mathbf{h}_i;\mathbf{e}_i'\odot \mathbf{h}_i;|\mathbf{e}_i'-\mathbf{h}_i|)+\mathbf{b}_s
		\label{eq:interract}
	\end{equation}
    where $;$ is the concatenation operation and $\mathbf{e}_i'$ is a transformation of $\mathbf{e}_i$ using:
	\begin{equation}
		\mathbf{e}_i'=\mathbf{W}_e(\mathbf{e}_i)+\mathbf{b}_e
		\label{eq:emap}
	\end{equation}
	This transformation is \textbf{essential and cannot be ignored} because that $\mathbf{e}$ and $\mathbf{h}$ are in different vector space. \textit{Otherwise the training will not converge}.
	
	We use cross entropy as the optimization objective:
	\begin{equation}
		\begin{split}
			\mathbf{y}^s_i&=softmax(\mathbf{h}^s_i)\\
			\mathcal{L}^{\mathrm{s}} &= -\sum^{1}_{i=0} \log P^{\mathrm{s}}(\mathbf{y}^t_i|\mathbf{h}^s_i)
		\end{split}
		\label{eq:nll_ls}
	\end{equation}
	
    For the self-supervised learning, we still employ the masked training strategy to conduct training, where we assign the label for random masking is $1$ (position with errors), and $0$ for those unchanged positions. Let $p^s_{x_t}=\mathbf{y}^s_t[1]$ be the self-supervised probability of error, then we also set up a threshold $\theta^s$ to conduct diagnose as well:
	\begin{eqnarray}
		error_t=\begin{cases}
			1, \ \ \ \text{if} \ \ p^s_{x_t}>=\theta^s
			\\
			0, \ \ \ otherwise
		\end{cases}
	\end{eqnarray}
	where $error_t=1$ means that token $x_t$ is not correct in sentence $X$. 
	
	Note that BERT parameters are \textbf{frozen} during the self-supervised learning procedure, therefore we only conduct optimization for a small group of parameters in Eq.~\ref{eq:interract} and Eq.~\ref{eq:emap}, which is a light-scale training stage.
	
	\subsection{Ensemble Detection Methods}
	Obvious, we can collect all the detected error positions ($\mathcal{I}^e_u$ and $\mathcal{I}^e_s$) by unsupervised and self-supervised detectors respectively, which we name it ensemble detection  operation.
	
	\begin{table*}[!t]
		\centering
		\small
		\resizebox{1.8\columnwidth}{!}{
			\begin{tabular}{l|l|ccc|ccc}
				\Xhline{3\arrayrulewidth}
				\multirow{2}{*}{\textbf{TestSet}} & \multirow{2}{*}{\textbf{Model}}  & \multicolumn{3}{c|}{\textbf{Detection}} & \multicolumn{3}{c}{\textbf{Correction}} \\ 
				\cline{3-8}
				& & \textsc{Prec.} & \textsc{Rec.} & \textsc{F1} & \textsc{Prec.} & \textsc{Rec.} & \textsc{F1} \\
				\hline
				SIGHAN13
				& \textbf{\underline{Supervised Methods}}\\
				& LMC \cite{xie-etal-2015-chinese} & 79.8 & 50.0 & 61.5 & 77.6 & 22.7 & 35.1  \\
				& Hybird \cite{DBLP:conf/emnlp/WangSLHZ18} & 54.0 & 69.3 & 60.7 & - & - & 52.1 \\
				&Confusionset \cite{DBLP:conf/acl/WangTZ19} & 66.8 &73.1 & 69.8 &71.5 &59.5 &69.9 \\
				& SpellGCN \cite{DBLP:conf/acl/ChengXCJWWCQ20} & 82.6 &88.9 &85.7&  98.4 &88.4 &93.1\\
				%& GAD \cite{DBLP:conf/acl/GuoNWZX21} & 85.8& 89.5 &87.6 &99.0 &88.6 &93.5 \\
				\cline{2-8}
				& \textbf{\underline{Unsupervised Methods}}\\
				& \textbf{uChecker} (Sec.\ref{sec:uchecker}) & 81.6 &  \textbf{93.0} &  \textbf{86.9} & 95.8 & 93.1 & 94.4 \\
				& \ \ \  w/o self-supervised detection  & 83.3 & 90.3 & 86.7 & 96.6 &  \textbf{93.2} &  \textbf{96.8}\\
				& \ \ \  w/o confusionset & \textbf{84.3} & 89.0 & 86.6 & 89.8 & 86.4 & 88.1\\
				\hline
				SIGHAN14
				& \textbf{\underline{Supervised Methods}}\\
				& LMC \cite{xie-etal-2015-chinese} & 56.4 & 34.8 & 43.0 & 71.1 & 50.2 & 58.8  \\
				& Hybird \cite{DBLP:conf/emnlp/WangSLHZ18} & 51.9 & 66.2 & 58.2 & - & - & 56.1 \\
				&Confusionset \cite{DBLP:conf/acl/WangTZ19} & 63.2 & 82.5 & 71.6 & 79.3 & 68.9 & 73.7 \\
				& SpellGCN \cite{DBLP:conf/acl/ChengXCJWWCQ20} & 83.6 & 78.6 & 81.0 & 97.2 & 76.4 & 85.5\\
				%& GAD \cite{DBLP:conf/acl/GuoNWZX21} &85.1 &80.9 &82.9 &98.0 &79.2 &87.6 \\
				\cline{2-8}
				& \textbf{\underline{Unsupervised Methods}}\\
				& \textbf{uChecker} (Sec.\ref{sec:uchecker}) & 75.9 & 73.3 & 74.6 & 91.7 & \textbf{84.9} &85.0 \\
				& \ \ \  w/o self-supervised detection  & 72.4 & 66.1 & 69.2 & 92.9& 81.4& \textbf{86.8}\\
				& \ \ \  w/o confusionset & 78.0 & 68.5 & 72.9 & 84.3 & 78.2 & 78.3\\
				\hline
				SIGHAN15
				& \textbf{\underline{Supervised Methods}}\\
				& LMC \cite{xie-etal-2015-chinese} & 56.4 & 34.8 & 43.0 & 71.1 & 50.2 & 58.8  \\
				& Hybird \cite{DBLP:conf/emnlp/WangSLHZ18} &56.6 &69.4 &62.3 &-& -& 57.1 \\
				&Confusionset \cite{DBLP:conf/acl/WangTZ19} & 66.8 & 73.1 & 69.8 & 71.5 & 59.5 & 69.9 \\
				& SpellGCN \cite{DBLP:conf/acl/ChengXCJWWCQ20} & 88.9 & 87.7 & 88.3 & 95.7 & 83.9 & 89.4\\
				& PLOME \cite{DBLP:conf/acl/LiuYYZW20} & 94.5 & 87.4 & 90.8 & 97.2 & 84.3 & 90.3\\
				%& GAD \cite{DBLP:conf/acl/GuoNWZX21} &88.6 &87.8 &88.2 &96.3 &84.6 &90.1 \\
				\cline{2-8}
				& \textbf{\underline{Unsupervised Methods}}\\
				& \textbf{uChecker} (Sec.\ref{sec:uchecker}) & 85.6 & 79.7 & 82.6 & 91.6 & \textbf{84.8} &88.1 \\
				& \ \ \  w/o self-supervised detection  & 75.8 & 71.3 & 73.5 & 92.6& 84.5&88.4\\
				& \ \ \  w/o confusionset & 87.4 & 75.9 & 81.2 & 84.6& 77.7&81.0\\
				\Xhline{3\arrayrulewidth}
		\end{tabular}}
		\caption{The character-level performance on both detection and correction level. *We notice that character-level detection performance of scrips from \citet{DBLP:conf/aclnut/HongYHLL19} and  \citet{DBLP:conf/acl/WangTZ19} are same. But the correction performance is different. Usually the scrip from \citet{DBLP:conf/acl/WangTZ19} is used to conduct the correction evaluation.}
		\label{tbl:res_char}
	\end{table*}
	
	\subsection{Confusionset-Guided Fine-Training}
	As mentioned in Section~\ref{sec:udec}, inspired by the random masking strategy in the pretraining stage of BERT, we tailor design a confusionset-guided random masking strategy where the target token $x_t$ will probably be replaced using its corresponding tokens in the confusionset $\mathcal{C}(x_t)$. The masking rate will also be adjusted slightly. Recently we find that confusionset-guided fine-training strategy has been deployed in some related works~\cite{DBLP:conf/acl/LiuYYZW20,DBLP:conf/acl/GuoNWZX21}. 
	
	After fine-training, we can use the new BERT model to conduct self-supervised/unsupervised detection and unsupervised correction.
	
	\section{Experimental Setup}
	\subsection{Settings}
	The core technical components of our proposed uChecker is a pre-trained Chinese BERT-base model \cite{DBLP:conf/naacl/DevlinCLT19}. The most important parameters in our framework are the two thresholds $\theta^u$ and $\theta^s$ and we set them to be $0.1$ and $0.4$ for unsupervised detection and supervised detection respectively.
	For the supervised error detection training, Adam optimizer \cite{DBLP:journals/corr/KingmaB14} is used to conduct the parameter learning and partial of the training dataset from SIGHAN series are employed as the trainset.
	
	\subsection{Datasets}
	
	The overall statistic information of the datasets used in our experiments are depicted in Table~\ref{tab:datasets}. As did in the previous works, we also conduct evaluation on those three datasets: SIGHAN13, SIGHAN14, and SIGHAN15~\cite{DBLP:conf/acl-sighan/TsengLCC15}\footnote{\url{http://ir.itc.ntnu.edu.tw/lre/sighan8csc.html}}.

	\begin{table*}[!t]
		%\small
		\centering
		\resizebox{1.9\columnwidth}{!}{
			\begin{tabular}{l|l|cccc|cccc}
				\Xhline{3\arrayrulewidth}
				\multirow{2}{*}{\textbf{TrainSet}} & \multirow{2}{*}{\textbf{Model}}  & \multicolumn{4}{c|}{\textbf{Detection}} & \multicolumn{4}{c}{\textbf{Correction}} \\ 
				\cline{3-10}
				& & \textsc{Acc.} & \textsc{Prec.} & \textsc{Rec.} & \textsc{F1} & \textsc{Acc.} & \textsc{Prec.} & \textsc{Rec.} & \textsc{F1} \\
				\hline
				SIGHAN13
				& \textbf{\underline{Supervised Methods}}\\
				&FASPell \cite{DBLP:conf/aclnut/HongYHLL19} & - & 76.2 & 63.2&  69.1& -& 73.1&  60.5&  66.2 \\
				& SpellGCN \cite{DBLP:conf/acl/ChengXCJWWCQ20} & - & 80.1 & 74.4 & 77.2 & - & 78.3 & 72.7 & 75.4 \\
				%& MLM-phonetics \cite{DBLP:conf/acl/ZhangPZWHSWW21} & - & 82.0 & 78.3 & 80.1& -& 79.5& 77.0 & 78.2\\
				%& PHMOSpell \cite{DBLP:conf/acl/HuangLJZCWX20}  & 77.1 & 99.5 & 76.8 & 86.7 & 75.4 & 99.5 & 75.1 & 85.6\\
				& DCN \cite{DBLP:conf/acl/WangCWWHL21}  & - & 86.8  & 79.6 & 83.0 & - & 84.7&  77.7&  81.0\\
				%& GAD \cite{DBLP:conf/acl/GuoNWZX21} & - & 85.7 & 79.5 & 82.5 &-& 84.9&  78.7 & 81.6 \\
				\cline{2-10}
				& \textbf{\underline{ Unsupervised Methods}}\\
				& \textbf{uChecker} (Sec.\ref{sec:uchecker}, Ours) & 73.4 & 75.4 & 73.4 & 74.4 &70.8& 72.6&  70.8 & 71.7\\ 
				& \ \ \  w/o self-supervised detection & 73.9 & 78.0 & 73.7 & 75.8 & 72.0& 75.9& 71.8 & 73.8\\
				& \ \ \  w/o confusionset & 73.5 & 78.2 & 73.3 & 75.7 & 65.5 & 69.4 & 65.1 & 67.2\\
				\hline
				SIGHAN14
				& \textbf{\underline{Supervised Methods}}\\
				&FASPell \cite{DBLP:conf/aclnut/HongYHLL19} & - & 61.0 &53.5 &57.0 &-&59.4& 52.0& 55.4 \\
				& SpellGCN \cite{DBLP:conf/acl/ChengXCJWWCQ20} & - & 65.1 &69.5 &67.2 &-&63.1 &67.2 &65.3 \\
				%& MLM-phonetics \cite{DBLP:conf/acl/ZhangPZWHSWW21} & - & 66.2& 73.8 &69.8&-& 64.2& 73.8& 68.7\\
				%& PHMOSpell \cite{DBLP:conf/acl/HuangLJZCWX20}  & 78.5 & 85.3 &67.6 &75.5& 76.9 &84.7& 64.3& 73.1\\
				& DCN \cite{DBLP:conf/acl/WangCWWHL21}  & - & 67.4 &70.4 &68.9 &-&65.8 &68.7 &67.2\\
				%& GAD \cite{DBLP:conf/acl/GuoNWZX21} & - & 66.6& 71.8 &69.1&-& 65.0& 70.1 &67.5 \\
				\cline{2-10}
				& \textbf{\underline{Unsupervised Methods}}\\
				& \textbf{uChecker} (Sec.\ref{sec:uchecker}, Ours) & 73.3 & 61.7 & 61.5 & 61.6 &71.3& 57.6&  57.5 & 57.6\\ 
				& \ \ \  w/o self-supervised detection & 68.4 & 55.3 & 52.1 & 53.7 &66.7& 51.6& 48.7 & 50.1\\
				& \ \ \  w/o confusionset & 72.5 & 62.3 & 57.3 & 59.7 & 58.3 & 52.9 & 48.7 & 50.7\\
				\hline
				SIGHAN15
				& \textbf{\underline{Supervised Methods}}\\
				&*FASPell \cite{DBLP:conf/aclnut/HongYHLL19} & 74.2 & 67.6 & 60.0 & 63.5 & 73.7 & 66.6 & 59.1 & 62.6 \\
				&*Confusionset \cite{DBLP:conf/acl/WangTZ19} & - & 66.8 & 73.1 & 69.8 & - & 71.5 & 59.5 & 64.9 \\
				&*SoftMask-BERT \cite{DBLP:conf/acl/ZhangHLL20} & 80.9 & 73.7 & 73.2 & 73.5 & 77.4 & 66.7 & 66.2 & 66.4 \\
				&*Chunk \cite{DBLP:conf/emnlp/BaoLW20} & 76.8 & 88.1 & 62.0 & 72.8 & 74.6 & 87.3 & 57.6 & 69.4 \\
				& SpellGCN \cite{DBLP:conf/acl/ChengXCJWWCQ20} & - & 74.8 & 80.7 & 77.7 & - & 72.1 &77.7 & 75.9 \\
				%& MLM-phonetics \cite{DBLP:conf/acl/ZhangPZWHSWW21} & - & 77.5 & 83.1 &80.2&-& 74.9 &80.2& 77.5\\
				%& PHMOSpell \cite{DBLP:conf/acl/HuangLJZCWX20}  & 82.6 &90.1 &72.7 &80.5 &80.9 &89.6 &69.2 &78.1\\
				& DCN \cite{DBLP:conf/acl/WangCWWHL21}  & - & 77.1 & 80.9 & 79.0 & - & 74.5 & 78.2 & 76.3\\
				%& GAD \cite{DBLP:conf/acl/GuoNWZX21} & - & 75.6 &  80.4& 77.9& - & 73.2 &77.8 & 75.4 \\
				\cline{2-10}
				& \textbf{\underline{Unsupervised Methods}}\\
				& \textbf{uChecker} (Sec.\ref{sec:uchecker}, Ours) & 82.2 & 75.4 & 72.0 & 73.7 &79.9& 70.6& 67.3&68.9\\
				& \ \ \  w/o self-supervised detection & 74.0 & 65.7 & 61.1 & 63.3 &72.6& 62.5& 58.1&60.2\\
				& \ \ \  w/o confusionset & 81.4 & 76.2 & 68.5 & 72.1 &76.5& 65.1& 58.5 & 61.6\\
				\Xhline{3\arrayrulewidth}
		\end{tabular}}
		\caption{The sentence-level performance on both detection and correction level. Evaluation script is from  \citet{DBLP:conf/aclnut/HongYHLL19}. \textbf{*} indicates the supervised methods which our unsupervised methods can outperform. } 
		\label{tbl:res_sent}
	\end{table*}
	
	\subsection{Comparison Methods}
	\label{sec:baselines}
	Considering that we did not notice some typical unsupervised methods with good results. Therefore, in this Section we introduce several classical and stage-of-the-art supervised approaches for comparisons.\\
	%\textbf{NTOU} employs n-gram language model with a reranking strategy to conduct prediction \cite{DBLP:conf/acl-sighan/TsengLCC15}.\\
	%\textbf{NCTU-NTUT} also uses CRF to conduct label dependency modeling \cite{DBLP:conf/acl-sighan/TsengLCC15}. \\
	\textbf{HanSpeller++} employs Hidden Markov Model with a reranking strategy to conduct the prediction \cite{DBLP:conf/acl-sighan/ZhangXHZC15}.\\
    \textbf{LMC} presents a model based on joint bi-gram and tri-gram LM and Chinese word segmentation~\cite{xie-etal-2015-chinese}. \\
	\textbf{Hybrid} utilizes LSTM-based seq2seq framework to conduct generation \cite{DBLP:conf/emnlp/WangSLHZ18} and \textbf{Confusionset} introduces a copy mechanism into seq2seq framework~\cite{DBLP:conf/acl/WangTZ19}.\\
	\textbf{FASPell} incorporates BERT into the seq2seq for better performance \cite{DBLP:conf/aclnut/HongYHLL19}.\\
	\textbf{SoftMask-BERT} firstly conducts error detection using a GRU-based model and then incorporating the predicted results with the BERT model using a soft-masked strategy \cite{DBLP:conf/acl/ZhangHLL20}. Note that the best results of \textbf{SoftMask-BERT} are obtained after pre-training on a large-scale dataset with 500M paired samples.\\
	\textbf{SpellGCN} proposes to incorporate phonological and visual similarity knowledge into language models via a specialized graph convolutional network~\cite{DBLP:conf/acl/ChengXCJWWCQ20}.\\
	\textbf{Chunk} proposes a chunk-based decoding method with global optimization to correct single character and multi-character word typos in a unified framework~\cite{DBLP:conf/emnlp/BaoLW20}.\\
	%\textbf{PHMOSpell} integrates the pinyin and glyph representations for Chinese characters from audio and visual modalities into a pre-trained language model by a well-designed adaptive gating mechanism~\cite{DBLP:conf/acl/HuangLJZCWX20}.\\
	\textbf{PLOME} also employs a confusionset to conduct training of BERT. Besides character prediction, PLOME also introduces pronunciation prediction to learn the misspelled knowledge on phonic level~\cite{DBLP:conf/acl/LiuYYZW20}.\\
	%\textbf{TtT} proposes to use a non-autoregressive text generation framework to regard CSC as a text generation task. CRF model is introducted to enhance the dependecy modeling among the labels~\cite{DBLP:conf/acl/Li020}.\\
	\textbf{DCN} generates the candidate Chinese characters via a Pinyin Enhanced Candidate Generator and then utilizes an attention-based network to model the dependencies between two adjacent Chinese characters~\cite{DBLP:conf/acl/WangCWWHL21}.\\
	%\textbf{GAD} learns the global relationship of the potential correct input characters and the candidates of potential error characters. Confusionset-guided fine-training is also used~\cite{DBLP:conf/acl/GuoNWZX21}.\\
	%\textbf{MLM-phonetics} integrates phonetic features into language model by leveraging the powerful pre-training and fine-tuning method and confusionset-guided fine-training is also used~\cite{DBLP:conf/acl/ZhangPZWHSWW21}.

	\begin{table*}[!t]
    \small
    \centering
    \resizebox{1.65\columnwidth}{!}{
    \begin{tabular}{l|c|cccc|cccc}
    \Xhline{3\arrayrulewidth}
     \multirow{2}{*}{\textbf{Parameter}} & \multirow{2}{*}{\textbf{Value}}  & \multicolumn{4}{c|}{\textbf{Detection}} & \multicolumn{4}{c}{\textbf{Correction}} \\ 
     \cline{3-10}
     & & \textsc{Acc.} & \textsc{Prec.} & \textsc{Rec.} & \textsc{F1} & \textsc{Acc.} & \textsc{Prec.} & \textsc{Rec.} & \textsc{F1} \\
    \hline
    $\theta^s$ & 0.2 & 76.3 & 99.0 & 76.4 & 86.3 & 65.5 & 98.9 & 65.4 & 78.7\\
    & 0.4 & 76.8 & 99.3 & 76.8 & 86.6 & 66.4 & 99.2 & 66.2 & \textbf{79.4}\\
    & 0.5 & 75.3 & 99.3 & 75.3 & 85.6 & 64.9 & 99.1 & 64.6 & 78.3\\
    & 0.6 & 72.2 & 99.5 & 71.9 & 83.4 & 62.5 & 99.4 & 62.0 & 76.4\\
    & 0.8 & 61.2 & 99.4 & 60.7 & 75.3 & 54.6 & 99.3 & 54.8 & 69.8\\
    \hline
    $\theta^u$  & 0.0001 & 21.9 & 99.1 & 20.2 &33.5 & 21.9 & 99.1 & 20.2 & 33.5\\
    & 0.01 & 44.3 & 99.6 & 43.2 & 60.2 & 41.9 & 99.6 & 41.7 & 57.8\\
    & 0.1 & 53.3 & 98.2 & 53.0 & 68.9 & 49.7 & 98.1 & 49.4 & \textbf{65.7}\\
    & 0.5 & 53.1 & 97.9 & 53.0 & 68.8 & 48.2 & 97.7 & 48.1 & 64.5\\
    & 0.9 & 44.2 & 96.7 & 44.3 & 60.8 & 38.4 & 96.2 & 38.4 & 54.9\\
    \Xhline{3\arrayrulewidth}
    \end{tabular}}
    \caption{Parameter tuning on devset of SIGHAN2015 (sentence-level). Due to the limited computing resource, we only conduct parameter tuning independently. Finally, we let $\theta^u=0.1$ and $\theta^s=0.4$.}
    \label{tbl:results_theta}
\end{table*}
	
	\subsection{Evaluation Metrics}
	Following the above mentioned works, we employ character-level and sentence-level \textbf{Accuracy}, \textbf{Precision}, \textbf{Recall}, and \textbf{F1-Measure} as the automatic metrics to evaluate the performance of all systems. Besides the official java-based evaluation toolkit (sentence-level)~\cite{DBLP:conf/acl-sighan/TsengLCC15}\footnote{\url{http://nlp.ee.ncu.edu.tw/resource/csc.html}}, as did in the previous works, we also report and compare the results evaluated by the tools from FASPell (character-level and sentence-level)~\cite{DBLP:conf/aclnut/HongYHLL19}\footnote{\url{https://github.com/iqiyi/FASPell}} and Confusionset (character-level)~\cite{DBLP:conf/acl/WangTZ19}\footnote{https://github.com/sunnyqiny/Confusionset-guided-Pointer-Networks-for-Chinese-Spelling-Check}.
	
	\section{Results and Discussions}
	
	\subsection{Main Results}
    \paragraph{Character-level Evaluation} Table~\ref{tbl:res_char} depicts the evaluation results on character-level on the datasets of SIGHAN13, SIGHAN14, and SIGHAN15. It is obvious most of the baseline methods are published recently and their performance are very strong. More importantly, almost all of the models are supervised learning based approaches and some of them are even trained using external large-scale datasets. Nevertheless, \textbf{surprisingly}, our proposed unsupervised framework uChecker has obtained comparable or even better results than those strong baseline methods. 
    Moreover, during the investigation about the evaluation methods, we notice that character-level detection performance of scrips from \citet{DBLP:conf/aclnut/HongYHLL19} and  \citet{DBLP:conf/acl/WangTZ19} are same. But the correction performance is different. Usually \citet{DBLP:conf/acl/WangTZ19} is used to conduct the correction evaluation and results reporting.

	\paragraph{Sentence-level Evaluation} Figure~\ref{tbl:res_sent} depicts the evaluation results in sentence-level on those three datasets. Evaluation script is employed from  \citet{DBLP:conf/aclnut/HongYHLL19} in order to align the results and to conduct comparing fairly. We also find that the official evaluation tool will output large values though the predicted results are same. We are trying to figure out the reasons.
	
	From Table~\ref{tbl:res_sent} we can observe that our  proposed unsupervised framework uChecker has obtained comparable or even better results than those strong baseline methods in the sentence-level as well. 
	
	\subsection{Parameter Tuning}
	Considering that the proposed unsupervised model uChecker is simple and straightforward, there are only two hyperparameters in our framework, $\theta^u$ and $\theta^s$, which are the threshold values to conduct spelling diagnosis for unsupervised detectors and supervised detectors respectively. And we only need to tune those two parameters. The tuning is conducted on the validation set of SIGHAN2015.
	Due to the limited computing resource, we only conduct tuning independently. Finally, we let $\theta^u=0.1$ and $\theta^s=0.4$.

	\subsection{Performance on small datasets}
	It is surprising that uChecker outperforms all the strong supervised  baselines on datasets SIGHAN13 in character-level evaluation, as shown in Figure~\ref{tbl:res_char}. After investigations we believe that the main reason is that the scale of trainset of SIGHAN13 (350) is much smaller than the other two corpora (6,526 and 3,174). This interesting phenomenon also  verify the advantages of the unsupervised learning based methods, especially for the task of CSC which is very difficult for collecting real labelled datasets.

	\subsection{Ablation Analysis}
	In the main results tables Table~\ref{tbl:res_char} and Table~\ref{tbl:res_sent}, we also provide the results of our model uChecker without the components of self-supervised detection and confusionset guided fine-training and decoding. Generally, the experimental results demonstrate that the corresponding components can indeed improve the performance. 
	
	\section{Conclusion}
	In this paper, we propose a framework named \textbf{uChecker} to conduct unsupervised spelling error detection and correction. Masked pretrained language models such as BERT are introduced as the backbone model. We also propose a confusionset-guided masking strategy to fine-train the model to further improve the performance. Experimental results on standard datasets demonstrate the effectiveness of our proposed model uChecker.
	%in terms of character-level and sentence-level Accuracy, Precision, Recall, and F1-Measure on tasks of spelling error detection and correction. 
	
    \section*{Acknowledgement}
	We thank the anonymous reviewers whose suggestions helped clarify this work. This research is supported by the National Natural Science Foundation of China (No.62106105), the CCF-Tencent Open Research Fund (No.RAGR20220122), the Scientific Research Starting Foundation of Nanjing University of Aeronautics and Astronautics (No.YQR21022), and the High Performance Computing Platform of Nanjing University of Aeronautics and Astronautics.
	
	% Entries for the entire Anthology, followed by custom entries
	\bibliography{anthology,custom}
	\bibliographystyle{acl_natbib}

\end{document}